\documentclass[final,1p,times,twocolumn]{elsarticle}

\usepackage{amssymb}
\usepackage{amsmath}
\usepackage{amsfonts}

\usepackage{lineno}
\usepackage{graphicx}
\usepackage{subfigure}
\usepackage{mathtools}
\usepackage{amsthm}
\usepackage{algorithm} 
\usepackage{algorithmic}  

\usepackage{hyperref}

\journal{ }

\begin{document}

\begin{frontmatter}

\title{AVATAR: \textbf{A}dversarial self-super\textbf{V}ised domain \textbf{A}daptation network for \textbf{TAR}get domain}



\author[1]{Jun Kataoka}
\ead{jkataok1@binghamton.edu}
\author[2]{Hyunsoo Yoon\corref{cor1}}
\cortext[cor1]{Corresponding author}
\ead{hs.yoon@yonsei.ac.kr}

\address[1]{The State University of New York at Binghamton, Department of Systems Science and Industrial Engineering, 
            Engineering Bldg, L2,
            13905,
            NY, United States}
\address[2]{Yonsei University, Department of Industrial Engineering, 
            50 Yonsei-ro,
            Seodaemun-gu,
            03722,
            Seoul,
            Republic of Korea}
            

\begin{abstract}
This paper presents an unsupervised domain adaptation (UDA) method for predicting unlabeled target domain data, specific to complex UDA tasks where the domain gap is significant. Mainstream UDA models aim to learn from both domains and improve target discrimination by utilizing labeled source domain data. However, the performance boost may be limited when the discrepancy between the source and target domains is large, or the target domain contains outliers. To explicitly address this issue, we propose the \textbf{A}dversarial self-super\textbf{V}ised domain \textbf{A}daptation network for the \textbf{TAR}get domain (AVATAR) algorithm. It outperforms state-of-the-art UDA models by concurrently reducing domain discrepancy while enhancing discrimination through domain adversarial learning, self-supervised learning, and sample selection strategy for the target domain, all guided by deep clustering. Our proposed model significantly outperforms state-of-the-art methods on three UDA benchmarks, and extensive ablation studies and experiments demonstrate the effectiveness of our approach for addressing complex UDA tasks.
\end{abstract}


\end{frontmatter}


\section{Introduction}\label{sec:introduction}
With the advancement of computer and sensing technology, large amounts of data have become available in various fields, such as manufacturing, healthcare, and transportation. However, the cost of acquiring labels for all this data remains extremely high. Accurate diagnosis or classification is crucial for the proper treatment of patients and maintaining the quality of manufacturing products. However, models based on limited or no labels often cannot achieve optimal results. Even when a sufficient number of labels exist for a specific domain, changes in the dataset are inevitable due to environmental factors. For example, process drift is a natural occurrence in manufacturing due to tool aging or incomplete maintenance. In the medical domain, cohort changes are inevitable due to the duration of data collection, changes in the acquisition protocol, or multiple data sources from various institutions. Since using the dataset from a single domain may not be sufficient for the target domain, leveraging labeled data from similar domains can significantly improve the learning process. This process, called domain adaptation, aims to utilize labeled data from one or more related domains to build models for a target domain.

In the Unsupervised Domain Adaptation (UDA) task, we are given labeled data from the source domain $\mathcal{S}$ and unlabeled data from the target domain $\mathcal{T}$; the objective is to adapt a model trained on $\mathcal{S}$ to $\mathcal{T}$. Generally, the UDA task assumes that different domains share identical conditional distributions of given input labels, but the marginal distributions of labels are different (e.g., label shift) \cite{quinonero-candela_when_2009}. To handle the label shift, classical UDA methods \cite{blitzer_learning_2007, mansour_domain_2009} adhere to a domain learning theory \cite{ben-david_analysis_2006}. The theory indicates that the target domain error can be minimized by bounding the source domain error and the domain discrepancy between the two domains \cite{zellinger_robust_2019}, leading to the idea of designing a representation function while minimizing the domain divergence and the classifier error \cite{blitzer_learning_2007, mansour_domain_2009}. As a result, various methods have been proposed for tackling the UDA task, including 1) domain adversarial learning \cite{csurka_domain-adversarial_2017, tzeng_adversarial_2017, hoffman_cycada_2018, zhang_hybrid_2020, zhang_transferable_2020}, 2) domain discrepancy minimization \cite{long_deep_2017, yan_weighted_2020}, 3) target domain pseudo-labeling or self-supervised learning \cite{zhang_adversarial_2021, na_fixbi_2021, li_pseudo-labeling_2023}, 4) deep clustering \cite{tang_unsupervised_2020, li_unsupervised_2022}, or combinations of these methods \cite{zhang_adversarial_2021}.

\begin{figure*}[!h]
\centering
\includegraphics[width=\textwidth]{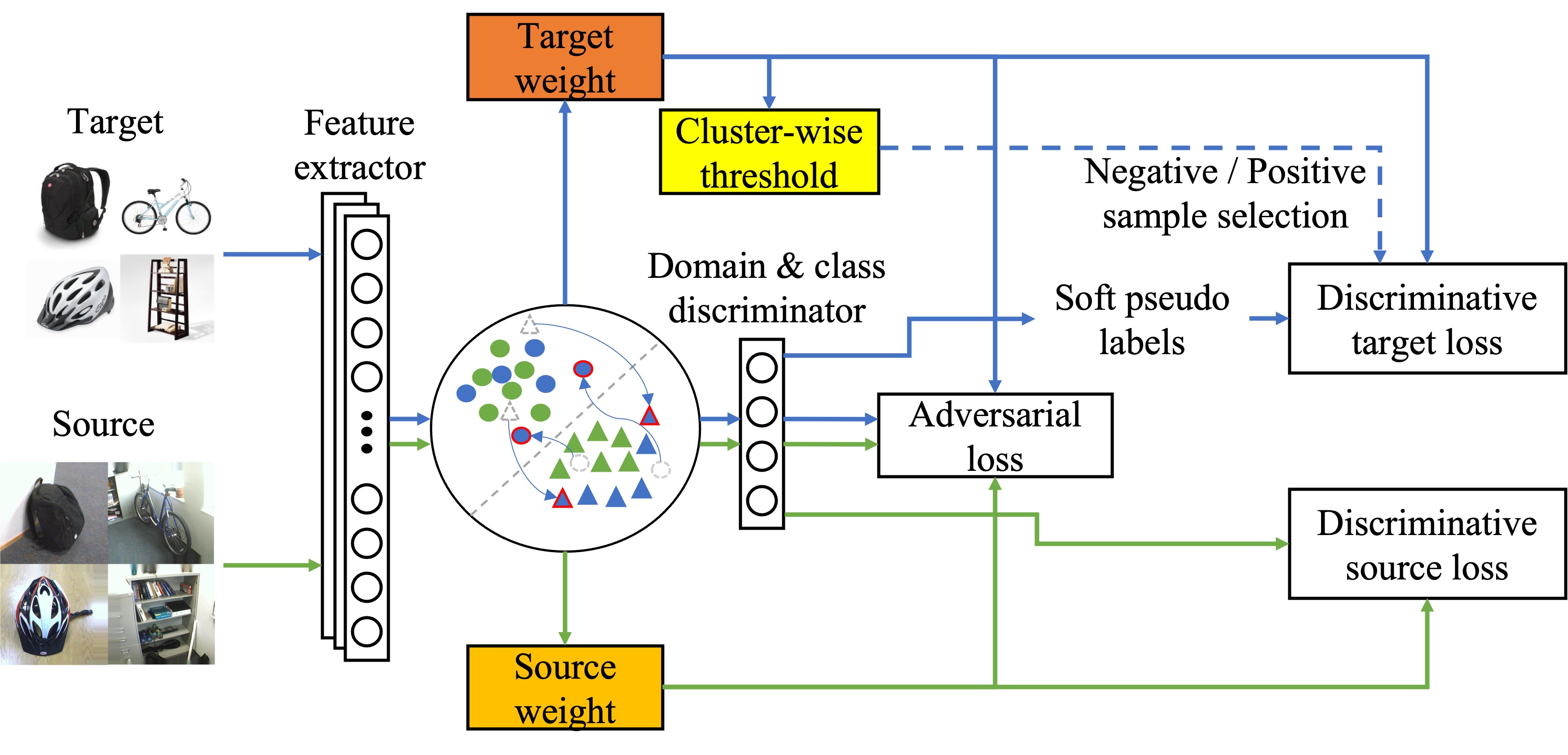}
\caption{The graphical illustration of the proposed AVATAR algorithm. Green and blue arrows indicate computational flow for the source and target domain, respectively. The extracted features using the feature extractor are embedded into a domain-invariant space. Source and target weights are computed based on the cosine similarity between target cluster centroids and each sample in the domain invariant space. A cluster-wise threshold is computed based on the target weight. Each target domain sample is assigned as a negative or positive transfer sample based on the threshold value and target weight.}
\label{fig:vis_abstract}
\end{figure*}

However, many previous methods do not explicitly consider the case when the difference between the two domains is significant, and they may not perform well when the UDA task is complex (i.e., a model trained using only source domain data cannot form good discriminative decision boundaries for the target domain). In this paper, we present the \textbf{A}dversarial self-super\textbf{V}ised domain \textbf{A}daptation network for \textbf{TAR}get domain (AVATAR) algorithm, which outperforms state-of-the-art UDA image classification models, especially for complex UDA tasks. In essence, the proposed model incrementally reduces the discrepancy between $\mathcal{S}$ and $\mathcal{T}$ using a domain adversarial learning, while also enhancing discrimination in $\mathcal{T}$ through a discriminative self-supervised learning and a sample selection strategy for the target domain, all guided by deep clustering. Unlike previous methods, AVATAR iteratively refines the discrimination boundaries in the domain-invariant feature space based on both cluster-wise weights and the sample selection strategy for the target domain to handle a large domain gap. The proposed model is the first of its kind to explicitly tackle complex UDA tasks, where a model trained on source domain data does not perform well (i.e., average accuracy less than 70\% on the target domain).  The graphical illustration of the proposed AVATAR algorithm is provided in Fig. \ref{fig:vis_abstract}.

We test AVATAR on three standard UDA benchmarks: Office-31 \cite{hu_unsupervised_2020}, Office-Home \cite{venkateswara_deep_2017}, and ImageCLEF \cite{10.1007/978-3-319-11382-1_18}. Our results show that our proposed model outperforms existing state-of-the-art methods on these three benchmarks, particularly in complex adaptation tasks. Additionally, we conduct extensive ablation studies on each component of AVATAR and find that our proposed approach significantly improves model performance. We also test AVATAR on imbalanced UDA tasks, where the label distribution of the target domain is imbalanced. The results suggest the proposed model's robustness for tackling the realistic UDA task in practice.

Overall, AVATAR's effectiveness in handling complex UDA tasks makes it a valuable tool for practical applications. Furthermore, with its potential to boost the performance of models in various domains in practice, AVATAR represents a significant contribution to the field of unsupervised domain adaptation.

\section{Related Work}
\label{relatedwork}
Adversarial learning-based domain adaptation has become a popular technique for reducing domain discrepancies by optimizing tailored loss functions in an adversarial manner. Originally inspired by the conceptual framework of the generative adversarial networks \cite{goodfellow_generative_2014}, various adversarial learning-based UDA methods have been proposed. The domain adversarial neural network (DANN) \cite{csurka_domain-adversarial_2017} is introduced as one of the first domain adversarial learning methods, which aims to learn a domain-invariant representation by back-propagating the reverse gradients of the domain classifier. The adversarial discriminative domain adaptation (ADDA) \cite{tzeng_adversarial_2017} extends and generalizes DANN, in which the source and target domain samples are separately trained with multiple networks to improve the target discrimination. Motivated by the Cycle-GAN \cite{zhu_unpaired_2017}, the cycle-consistent adversarial domain adaptation (CyCADA) \cite{hoffman_cycada_2018} uses a cycle-consistency loss within the adversarial learning framework to capture semantic consistency between two domains. Furthermore, Saito et al. \cite{saito_maximum_2018} introduces a domain adversarial approach to align the distribution of the source and target domains while utilizing the task-specific decision boundaries.

Deep clustering-based domain adaptation utilizes conditional entropy minimization \cite{grandvalet_semi-supervised_2004} to minimize the within-class distance in the shared feature space between the domains. In the article by Xie et al. \cite{xie_unsupervised_2016}, they introduced a technique called deep embedded clustering (DEC), which involves using deep neural networks to learn feature representations and cluster assignments simultaneously. DEC achieves this by mapping data from its original space to a lower-dimensional feature space and then iteratively optimizing a clustering objective in that space. The work of Kang et al. \cite{kang_contrastive_2019} uses a contrastive learning-based loss function within its deep clustering framework to minimize the within-class and between-class domain discrepancies simultaneously. Recently, structurally regularized deep clustering (SRDC) \cite{tang_unsupervised_2020} has been proposed to improve the target discrimination of clustering by structurally regularizing source samples. Contrary to other studies based on conditional entropy minimization, which explicitly forces the alignment between different domains, SRDC focuses on intrinsic target discrimination without explicitly aligning domains by utilizing a weighting scheme based on the K-means algorithm.

Using self-supervised learning based on model prediction or clustering is another standard technique used to address UDA tasks. Typically, a pre-trained source classifier assigns the initial target pseudo-labels. Subsequently, different algorithms refine the target pseudo-labeling by promoting the distribution alignment between domains. Moving semantic transfer network (MSTN) \cite{xie_learning_2018} combines semantic matching and domain adversarial loss to obtain pseudo-labels. Incremental collaborative and adversarial networks (iCAN) \cite{zhang_collaborative_2018} utilizes the domain adversarial learning, but it also expands the training set and selects samples to generate pseudo-labels interactively. Also, the adversarial continuous learning in unsupervised domain adaptation (ACDA) \cite{zhang_adversarial_2021} integrated confident continuous learning into adversarial learning, increasing the model robustness by incorporating a high-confidence sample from the target domains. In a similar line of work, \cite{gu_spherical_2020} proposes a robust pseudo-label loss in the spherical feature space. The proposed spherical adversarial training method weighs the target domain samples by Gaussian-uniform model in the spherical feature space.

Notably, many of these previous methods emphasize improving the performance on general UDA benchmarks, and considering the negative effects of outliers or difficult-to-classify samples in the target domain is often out of their main scopes. Recently, the discriminative adversarial domain adaptation (DADA) \cite{tang_discriminative_2020} introduced a mutually inhibitory relationship between class category and domain predictions to promote the joint distribution alignment across domains. Extension of DADA to partial and open-set domain adaptation is also proposed to handle outliers in source and target domains by category-level weighting mechanism. However, the proposed model's aim for the UDA task is still limited to concurrently considering both domain and class predictions. Also, FixBi \cite{na_fixbi_2021} utilizes bidirectional matching with confidence-based pseudo-labeling strategies to increase model robustness by selecting high-confidence target samples to generate pseudo-labels while penalizing low-confident target samples for prediction.

Though AVATAR is inspired by these previous methods in terms of effectively utilizing the mutually inhibitory relationship, outliers in the target domain, and category-level weighting scheme, it explicitly puts more focus on solving complex UDA transfer tasks. Each of the buildings of AVATAR is strongly tailored for generating more robust discriminative boundaries, especially when the domain gap is significant.

\section{Method}
In this section, we explain each building block of AVATAR. We consider data from the labeled source domain as $\mathcal{X}^s = \{(x_i^s,y_i^s)\}_{i=1}^{N^s} \in \mathcal{S}$ and the unlabeled target domain data as $\mathcal{X}^t = \{(x_i^t)\}_{i=1}^{N^t} \in \mathcal{T}$, where $N^s$ and $N^t$ denote the sample sizes of the data.

\subsection{Clustering-guided weighted domain adversarial learning}\label{sub:adversarial}
Previous studies have proposed various frameworks to reduce the domain discrepancy between $\mathcal{S}$ and $\mathcal{T}$, and they often employ the domain adversarial training between the feature extractor and the domain classifier \cite{csurka_domain-adversarial_2017}. Among these methods, \cite{liu_adversarial_2021, tran_gotta_2019, salimans_improved_2016, tang_discriminative_2020} suggest forcing the domain classifier to be aware of the classification result enhances the generation of the domain-invariant features while improving the target discrimination. Following this notion, we design a joint discriminator and classifier $g$ and enforce it to be implicitly aware of the classification result while adversely trained with a feature extractor $f$. 

Here, we denote the mapping from the input to the latent feature space by the feature extractor as $f: \mathcal{X}^s, \mathcal{X}^t \mapsto \mathcal{Z}$, and the outputs from the feature extractor $f$ are defined as $\mathbf{z}_i^{s} = f(x_i^s; \theta)$ and $\mathbf{z}_i^{t} = f(x_i^t; \theta)$, where $\theta$ is a set of network parameters for the feature extractor. Similarly, the mapping by the joint discriminator and classifier is defined as $g: \mathcal{Z} \mapsto \mathcal{P}$, and the source and target probability vectors in $\mathcal{P}$ after the softmax operation are denoted as $\mathbf{p}_i^{s} = g(\mathbf{z}_i^{s}; \vartheta)$ and $\mathbf{p}_i^{t} = g(\mathbf{z}_i^{t}; \vartheta)$, where $\vartheta$ is a set of network parameters for $g$. Similar to \cite{tang_discriminative_2020}, we add an additional output unit to represent the domain to which the input belongs to inform the joint discriminator and classifier $g$ of the class and domain predictions. In other words, the output of $g$ has $K+1$ dimensions consisting of predictions for $K$ classes and a domain prediction. Therefore, $|\mathcal{P}|=K+1$ holds for the $K$ class classification tasks. Consequently, we have $p^t_{i, k}$ and $p^s_{i, k}$ for $k=1, \dots, K+1$ as the $k^{th}$ elements of $\mathbf{p}_i^s$ and $\mathbf{p}_i^t$, where the $(K+1)^{th}$ element represents whether an input belongs to the source domain $\mathcal{S}$ or the target domain $\mathcal{T}$.

In previously proposed domain adversarial learning methods \cite{zhang_adversarial_2021, gu_spherical_2020}, the domain adversarial learning is gradually carried on as continuous learning, bringing high-confidence transfer examples (i.e., positive samples) from the target domain to the source domain based on a probability threshold. In our proposed method, however, we intend to align all the target domain samples, including the low-confident transfer examples (i.e., negative samples) to utilize all the target samples during the training. This notion is based on that previously proposed continuous adversarial learning methods \cite{zhang_adversarial_2021, gu_spherical_2020} may discard many target samples for a complex UDA task where the domain discrepancy between source and target is significant. 

In a complex UDA task, many target samples may belong to the negative samples in the earlier stage of the training, and generating domain-invariant features based on a few positive target samples may damage the target discrimination during the training. Also, using the prediction probabilities to weigh and select target domain samples inherently assumes the prediction output is correct. However, such an assumption could be violated when dealing with a complex UDA task where model prediction could be wrong for difficult-to-classify or outlier samples. Therefore, instead of utilizing prediction probability, our proposed framework utilizes an iterative clustering method (e.g., K-means) to assign sample weights to the target domain samples. Furthermore, since the iterative clustering method is independent of a neural network's training process, our proposed weighting scheme demonstrates greater robustness in the face of outliers and is more effective in addressing UDA tasks where accurate prediction for target domain samples proves to be challenging.

To this end, we gradually align all the source and target samples based on their distances from the corresponding target class centroids in $\mathcal{Z}$ estimated using the kernel K-means algorithm. Specifically, we compute the target class centroids $\{\mathbf{c}_k^{t}\}_{k=1}^K \in \mathcal{Z}$ based on the kernel K-means algorithm at the beginning of each epoch. The weights for each instance from both domains are computed using cosine similarities \cite{xie_unsupervised_2016}

\begin{equation} \label{eq1}
\{w_i^s\}_{i=1}^{N^s} = 0.5(1+ \frac{{\mathbf{c}^{t}_{k={y^s_i}}}^T \mathbf{z}_i^{s}}
                                  {\|\mathbf{c}^{t}_{k={y^s_i}}\| \|\mathbf{z}_i^{s}\|}) \in [0, 1],
\end{equation}
and
\begin{equation} \label{eq2}
\{w_i^t\}_{i=1}^{N^t} = 0.5(1+ \frac{{\mathbf{c}^{t}_{k={\hat{y}^t_i}}}^T \mathbf{z}_i^{t}}
{\|\mathbf{c}^{t}_{k={\hat{y}^t}_{i}}\| \|\mathbf{z}_i^{t}\|}) \in [0, 1],
\end{equation}

where $y^s_i$ is the class label for the $i^{th}$ sample in the source domain, $\hat{y}^t_i$ is the predicted class label for the $i^{th}$ target domain sample using kernel K-means algorithm.
Note that $\mathbf{c}^{t}_{k={\hat{y}^t_i}}$ denotes the target class centroid in $\mathcal{Z}$ for the $k^{\text{th}}$ predicted target class, and $\{w_i^s\}_{i=1}^{N^s}$ and $\{w_i^t\}_{i=1}^{N^t}$ are computed based on the distance from the target class centroid $\mathbf{c}^{t}_{k={\hat{y}^t_i}}$. 

Using $\{w_i^s\}_{i=1}^{N^s}$ and $\{w_i^t\}_{i=1}^{N^t}$, we weigh and prioritize samples closer to their corresponding target class centroids within a domain adversarial learning framework, while mitigating the effects of samples farther away from the target class centroids. By including this weighting mechanism, we can mitigate the effects of instances that are farther away from the target class centroids and focus on aligning the features of instances that are more likely to be similar across domains. Therefore, our proposed model can generate a more robust and discriminative domain-invariant representation between $\mathcal{S}$ and $\mathcal{T}$. 

We first fix $f$ and train $g$ to classify domains based on the features from both domains using cross-entropy, yielding
\begin{equation} \label{eq3}
\min_{g} L_{adv_{g}} = - \frac{1}{N_s} \sum_{i=1}^{N_s} w_i^s \log{p_{i, K+1}^s}
-\frac{1}{N_t} \sum_{i=1}^{N_t} w_i^t \log{(1-p_{i, K+1}^t)}.
\end{equation}
Then, we fix $g$ and let $f$ learn feature representations such that all instances are classified to wrong domains by $g$, yielding
\begin{equation} \label{eq4}
\min_{f} L_{adv_{f}} = - \frac{1}{N_s} \sum_{i=1}^{N_s} w_i^s \log{(1-p_{i, K+1}^s)}
-\frac{1}{N_t} \sum_{i=1}^{N_t} w_i^t \log{p_{i, K+1}^t}.
\end{equation}

By alternatively minimizing Eq.\eqref{eq3} and Eq.\eqref{eq4} using a gradient-descent-based algorithm, we aim to obtain domain-invariant representations between $\mathcal{S}$ and $\mathcal{T}$. Although information about class prediction is not directly included in Eq.\eqref{eq3} or Eq.\eqref{eq4}, $g$ is still implicitly aware of the class probability from the other $K$ output units, and the magnitude of the loss is determined based on the values of $w_i^s$ and $w_i^t$. This way, the model can gradually learn domain-invariant features that are robust even if the initial domain discrepancy is large between the source and target domains.

\subsection{Sample selection strategy for the target domain guided by within-cluster distance}\label{sub:selection}
In our domain adversarial learning framework, source and target domain samples are aligned based on cosine similarity-based weights $\{w_i^s\}_{i=1}^{N^s}$ and $\{w_i^t\}_{i=1}^{N^t}$. In other words, gradually reducing the within-cluster distance by forming tighter clusters would push these weights to their upper bounds, enabling the model to effectively utilize all the samples to generate the domain-invariant features during model training. Furthermore, although previous deep clustering-based domain adaptation methods such as \cite{tang_unsupervised_2020, xie_unsupervised_2016} utilize the idea of creating discriminative clusters for the target domain, our proposed method extends this idea by creating clusters for the positive transfer samples, while still utilizing as much information as possible from the negative transfer samples from the target domain.

To determine negative and positive target samples, we compute the within-cluster mean of $\{w_i^{t}\}_{i=1}^{N^t}$ for the $k^{th}$ class in the target domain as:
\begin{equation} \label{eq7}
  \bar{w}_k^t = \frac{1}{N_{k=\hat{y}_{i}^t}^t}\sum_{i=1}^{N^t} I[k=\hat{y}_i^t] w_i^{t},
\end{equation}
where $N_{k=\hat{y}_{i}^t}^t$ denotes the number of target samples predicted as $\hat{y}_i^t$ by kernel K-means algorithm.
We then define a confidence threshold $\tau_k^t$ for each cluster to evaluate the confidence level of the target sample in the $k^{th}$ cluster as follows:
\begin{equation} \label{eq8}   \tau_k^t = \bar{w}_k^t - \sqrt{\frac{1}{N_{k=\hat{y}_{i}^t}^t}\sum_{i=1}^{N^t} I[k=\hat{y}_i^t] (w_i^{t})^2 - (\bar{w}_k^t)^2},
\end{equation}
where $I[k=\hat{y}_i^t]$ is an indicator function that is 1 when the sample $i$ belongs to the $k^{th}$ cluster, otherwise it is 0.
Note that the second term in Eq.\eqref{eq8} represents the within-cluster deviation of $\{w_i^{t}\}_{i=1}^{N^t}$ for the $k^{th}$ cluster, and $\tau_k^t$ is determined based on this deviation. Once $\{\tau_k^t\}_{k=1}^K$ is calculated, we assign target samples classified as the $k^{th}$ cluster as positive if $w_i^{t} \geq \tau_k$ and negative if $w_i^{t}<\tau_k$. Contrarily to the previous methods that use a global threshold for all the target domain samples \cite{na_fixbi_2021, gu_spherical_2020}, our proposed method determines the level of confidence for each target sample based on the within-cluster distance without using hyper-parameters, resulting in more robust and accurate detection of negative target samples.

\subsection{Weighted self-supervised learning for the target domain} \label{sub:clustering}

We further integrate the sample selection strategy for the target domain into the self-supervised learning framework. We first normalize the softmax output $\mathbf{p}_i^s$ and $\mathbf{p}_i^t$ by dividing each class probability by the sum of all class probabilities, resulting in $\mathbf{p}_i^{s'}, \mathbf{p}_i^{t'} \in [ 0, 1 ] $. We then define a target pseudo-label distribution, denoted as $\mathcal{Q}^t$, as an auxiliary counterpart of $\mathcal{P}^{t'}$ which is the normalized class probability space. 
For $\mathcal{P}^{t'}$, we minimize the KL divergence between $\mathcal{Q}^t$ and $\mathcal{P}^{t'}$ while maximizing the entropy of $\mathcal{Q}^t$ \cite{dizaji_deep_2017, tang_unsupervised_2020}. By setting the approximated gradient as zero for this optimization problem, a closed-form update rule for $\mathcal{Q}^t$ is obtained as follows:
\begin{equation}\label{eq5}
  q_{i, k}^t = \frac{{p}_{i, k}^{t'}
                            (\sum_{i'=1}^{N_t} {p}_{i', k}^{t'})^{0.5}}
                    {\sum_{k'=1}^K {p}_{i, k'}^{t'} /
                      (\sum_{i'=1}^{N_t} {p}_{i', k'}^{t'})^{0.5}}.
\end{equation}

Once $\mathcal{Q}^t$ is updated, we simply update $f$ and $g$ by minimizing the cross-entropy loss using $\mathcal{Q}^t$ as a soft-pseudo label, which yields to
\begin{equation} \label{eq6}
  \min_{f, g} L_{dis}^t = - \frac{1}{N_t} \sum_{i=1}^{N_t} w_i^{t}
  \sum_{k=1}^K q_{i, k}^t \log{{p}_{i,k}^{t'}}.
\end{equation}

As in Eq.\eqref{eq3} and Eq.\eqref{eq4}, this loss function is designed to prioritize target samples that are closer to their assigned $k^{th}$ target cluster centroid $\mathbf{c}^{t}_{k={\hat{y}^t}_{i}}$ by utilizing the instance weight $w_i^{t}$. Sharing the same weights between the domain adversarial learning and self-supervised learning would gradually tighten each cluster to form robust target discrimination boundaries in the domain-invariant feature space, resulting to better performance when tackling complex UDA tasks where the target discrimination is difficult in the early stage of training.

For negative transfer samples, we aim to decrease the probability value for the $k^{th}$ class and increase the values for other classes. Additionally, we prioritize negative samples that are farther away from their target cluster centroids based on the value of $(1-w_i^{t})$, resulting in negative target domain samples located further away from the centroids being assigned larger weights and moved away from the cluster centroids.

As a result, we modify Eq.\eqref{eq6} as
\begin{equation} \label{eq9}
  L_{dis}^{t} = - \frac{1}{N_t}
   \sum_{i=1}^{N_t} \biggl [
    w_i^{t} \sum_{k=1}^K I[w_{i}^{t}\ \geq \tau_k] q_{i, k}^t \log{{p}_{i,k}^{t'}}
    + (1-w_i^{t}) \sum_{k=1}^K I[w_{i}^{t} < \tau_k] q_{i, k}^t \log{(1-{p}_{i,k}^{t'})}\biggl ].
\end{equation}

In many deep clustering methods for unsupervised domain adaptation (UDA), samples that are assigned to the same cluster centroid are often assumed to be of the same class, even if some samples are located far away from the centroids \cite{kang_contrastive_2019, pan_transferrable_2019, shu_dirt-t_2018, dizaji_deep_2017}. Such an assumption may become problematic for solving a complex UDA task, where negative target samples may potentially belong to other classes. Including all the target samples in the ordinary cross-entropy loss would negatively impact the model's target discrimination performance, even if the magnitude of gradients from them is controlled by instance weighting methods. Intuitively, Eq.\eqref{eq9} pushes away difficult-to-classify or outliers in the target domain further away from corresponding cluster centroids based on their magnitudes. Therefore, the purity of the resulting clusters for target domain samples is kept higher than the previous deep clustering methods for domain adaptation \cite{kang_contrastive_2019, pan_transferrable_2019, shu_dirt-t_2018, dizaji_deep_2017}, resulting in the robust formation of target discrimination boundaries, and efficient feature alignment through the domain adversarial training becomes possible.

To utilize labeled source domain samples, we utilize the discriminative source loss as a structural regularization \cite{tang_unsupervised_2020} through the cross-entropy loss weighted by $w_i^s$, resulting in the loss for each sample is proportional to its distance from the target cluster centroid. Our discriminative loss for the source domain is defined as:
\begin{equation} \label{eq10}
L_{dis}^s = - \frac{1}{N_s} \sum_{i=1}^{N_s} w_i^{s}
\sum_{k=1}^K I[k=y_i^s]\log{{p}_{i,k}^{s'}}.
\end{equation}
This loss is applied to the source samples which have labeled data, and it aims to ensure that the representation of the source domain is discriminative.

We combine Eq.\eqref{eq9} and Eq.\eqref{eq10} to obtain the overall discriminative loss function, defined as
\begin{equation} \label{eq11}
  L_{dis} = L_{dis}^t + L_{dis}^s.
\end{equation}
Finally, we have the overall objective of the AVATAR algorithm for $f$ and $g$ by minimizing Eq.\eqref{eq11} while optimizing Eq.\eqref{eq3} and Eq.\eqref{eq4} alternatively, yielding to

\begin{equation} \label{eq12}
  \min_{f} \lambda (L_{dis} + L_{adv_f}),
\end{equation}
and
\begin{equation} \label{eq13}
  \min_{g} \lambda (L_{dis} + L_{adv_g}),
\end{equation}
where $\lambda$ is a regularization parameter which controls the magnitude of discriminative and adversarial losses.

For further reference, the training procedure of AVATAR is provided in Algorithm \ref{alg:the_alg}.

\section{Experiments and Discussion}
In this section, we evaluate our proposed method by comparing it with state-of-the-art domain adaptation methods on three domain adaptation benchmarks: Office-31 \cite{hu_unsupervised_2020}, Office-Home \cite{venkateswara_deep_2017}, and ImageCLEF \cite{10.1007/978-3-319-11382-1_18}. Additionally, we validate the contribution of the proposed model through extensive ablation studies. We further evaluate the mechanism of AVATAR by examining its feature representations and confidence thresholds for the sample selection strategy. Finally, we test our proposed method for more realistic cases of domain adaptation, where the source and target domains' label distributions differ. Specifically, we test the proposed model's performance using imabalanced datsets created using the VISDA-2017 dataset.

\subsection{Dataset}

{\bf Office-31} \cite{hu_unsupervised_2020} is a widely used benchmark for evaluating domain adaptation in real-world scenarios. It includes three domains: Amazon (A), Webcam (W), and DSLR (D). The dataset comprises images from Amazon.com and office environments captured from different angles and lighting conditions using webcam or DSLR cameras. In total, it includes 4,110 images, with 31 categories per domain. {\bf Office-Home} is a challenging benchmark, which includes 15,000 images from four distinct domains: artistic images (Ar), clip art (Cl), product images (Pr), and real-world images (Rw). Each domain comprises of 65 different classes of images that pertain to the office and home equipment, as reported in \cite{venkateswara_deep_2017}. {\bf 
 ImageCLEF} \cite{10.1007/978-3-319-11382-1_18} is another popular domain adaptation benchmark. It consists of three domains derived from three public datasets: Caltech (C), ImageNet (I), and Pascal (P). Each domain contains 12 categories, and each category has 50 images.

\subsection{Implementation and Model Setup} \label{subsection:setup}
For the model training procedure, we follow the standard protocol for UDA \cite{csurka_domain-adversarial_2017, liu_adversarial_2021, long_conditional_2018} using all labeled source and unlabeled target data for training. In addition, as a baseline model architecture, we use the Image-Net-trained ResNet50 \cite{he_deep_2016} and replace the last fully connected layer with a task-specific fully connected one.
We first train the baseline model using only the source domain dataset as our initial model. We then fine-tune the model using both source and target domains in order to improve the model's performance on the target domain.
Our optimization method uses mini-batch Stochastic Gradient Descent with a dynamic learning rate. The learning rate is adjusted according to a schedule that changes as the training progresses \cite{long_learning_2015}. The schedule includes a decay factor based on the current training epoch, normalized between $[0, 1]$. Additionally, the task-specific layer's learning rate is 10 times higher than the other layers. To set the regularization parameter $\lambda$, we follow a similar approach as Long et al. \cite{long_learning_2015}, which involves increasing the value of $\lambda$ as the training progresses, using an exponential function. We train our model for 200 epochs using mini-batch sizes of 64. Additionally, we use the first five epochs as a warm-up period before applying our sample selection strategy for the target domain. For more information on the implementation details, our PyTorch code is available at \href{https://github.com/junkataoka/AVATAR}{https://github.com/junkataoka/AVATAR.}

\begin{table*}[!h]
\caption{Accuracy$(\%)$ on Office-31 for unsupervised domain adaptation (ResNet-50). The best accuracy is indicated in bold. \\ * Reproduced by \cite{gu_spherical_2020}}
\begin{center}
\resizebox{\textwidth}{!}{
 \begin{tabular}{|c|c|c|c|c|c|c|c|}
 \hline
 Method & A $\rightarrow$ W & D $\rightarrow$ W & W $\rightarrow$ D & A $\rightarrow$ D & D $\rightarrow$ A & W $\rightarrow$ A & Avg \\ [0.5ex]
 \hline\hline
 ResNet-50\cite{he_deep_2016} & 68.4$\pm$0.2 & 96.7$\pm$0.1 & 99.3$\pm$0.1 & 68.9$\pm$0.2 & 62.5$\pm$0.3 & 60.7$\pm$0.3 & 76.1 \\
 DANN\cite{long_learning_2015} & 82.0$\pm$0.4 & 96.9$\pm$0.2 & 99.1$\pm$0.1 & 79.7$\pm$0.4 & 68.2$\pm$0.4 & 67.4$\pm$0.5 & 82.2 \\
 MSTN*\cite{xie_learning_2018}& 91.3 & 98.9 & {\bf 100.0} & 90.4 & 72.7 & 65.6 & 86.5 \\
 CDAN+E\cite{long_conditional_2018}& 94.1$\pm$0.1 & 98.6$\pm$0.1 & {\bf 100.0}$\pm$0.0 & 92.9$\pm$0.2 & 71.0$\pm$0.3 & 69.3$\pm$0.3 & 87.7 \\
 SymNets\cite{zhang_domain-symmetric_2019} & 90.8$\pm$0.1 & 98.8$\pm$0.3 & {\bf 100.0}$\pm$0.0 & 93.9$\pm$0.5 & 74.6$\pm$0.6 & 72.5$\pm$0.5 & 88.4 \\
 GSDA\cite{hu_unsupervised_2020} & 95.7 & 99.1 & {\bf 100.0} & 94.8 & 73.5 & 74.9 & 89.7 \\
 CAN\cite{cui_gradually_2020} & 94.5$\pm$0.3 & 99.1$\pm$0.2 & 99.8$\pm$0.2 & 95.0$\pm$0.3 & 78.0$\pm$0.3 & 77.0$\pm$0.3 & 90.6 \\
 SRDC\cite{tang_unsupervised_2020} & 95.7$\pm$0.2 & 99.2$\pm$0.1 & {\bf 100.0}$\pm$0.0 & {\bf 95.8}$\pm$0.2 & 76.7$\pm$0.3 & 77.1$\pm$0.1 & 90.8 \\
 RSDA-MSTN\cite{gu_spherical_2020} & {\bf 96.1}$\pm$0.2 & {\bf 99.3}$\pm$0.2 & {\bf 100.0}$\pm$0.0 & {\bf 95.8}$\pm$0.3 & 77.4$\pm$0.8 & 78.9$\pm$0.3 & 91.1 \\
 FixBi\cite{na_fixbi_2021} & {\bf 96.1}$\pm$0.2 & {\bf 99.3}$\pm$0.2 & {\bf 100.0}$\pm$0.0 & 95.0$\pm$0.4 & 78.7$\pm$0.5 & 79.4$\pm$0.3 & 91.4 \\
 \hline
 AVATAR & 95.1$\pm$0.3 & 98.1$\pm$0.1& {\bf 100}$\pm$0.0 & 92.0$\pm$0.5 & {\bf 86.2}$\pm$0.2 & {\bf 85.5}$\pm$0.6 & {\bf 92.8}$\pm$0.1 \\
 \hline
 \end{tabular}
}
\label{table:1}
\end{center}
\end{table*}

\begin{table*}[!h]
\caption{Accuracy$(\%)$ on Office-Home for unsupervised domain adaptation (ResNet-50). The best accuracy is indicated in bold. \\ * Reproduced by\cite{gu_spherical_2020}}
\begin{center}
\resizebox{1.0\textwidth}{!}{
     \begin{tabular}{|c|c|c|c|c|c|c|c|c|c|c|c|c|c|}
     \hline
       Method&Ar$\rightarrow$Cl& Ar$\rightarrow$Pr & Ar$\rightarrow$Rw & Cl$\rightarrow$Ar & Cl$\rightarrow$Pr & Cl$\rightarrow$Rw & Pr$\rightarrow$Ar & Pr$\rightarrow$Cl & Pr$\rightarrow$Rw & Rw$\rightarrow$Ar & Rw$\rightarrow$Cl & Rw$\rightarrow$Pr & Avg \\ [0.5ex]
     \hline\hline
     ResNet-50\cite{he_deep_2016} & 34.9 & 50.0 & 58.0 & 37.4 &
       41.9 & 46.2 & 46.2 & 38.5 & 31.2 &
       60.4 & 53.9 & 41.2 & 46.1 \\
     DANN\cite{long_learning_2015} & 45.6 & 59.3 & 70.1 & 47.0 &
       58.5 & 60.9 & 46.1 & 43.7 & 68.5 &
       63.2 & 51.8 & 76.8 & 57.6 \\
     CDAN\cite{long_conditional_2018} & 49.0 & 69.3 & 74.5 & 54.4 &
       66.0 & 68.4 & 55.6 & 48.3 & 75.9 &
       68.4 & 55.4 & 80.5 & 63.8 \\
     MSTN*\cite{xie_learning_2018} & 49.8 & 70.3 & 76.3 & 60.4 &
       68.5 & 69.6 & 61.4 & 48.9 & 75.7 &
       70.9 & 55.0 & 81.1 & 65.7 \\
     SymNets\cite{zhang_domain-symmetric_2019} & 47.7 & 72.9 & 78.5 &
       64.2 & 71.3 & 74.2 & 63.6 & 47.6 & 79.4 &
       73.8 & 50.8 & 82.6 & 67.2 \\
     GSDA\cite{hu_unsupervised_2020} & 61.3 & 76.1 & 79.4 & 65.4 &
       73.3 & 74.3 & 65.0 & 53.2 & 80.0 &
       72.2 & 60.6 & 83.1 & 70.3 \\
     GVB-GD\cite{cui_gradually_2020} & 57.0 & 74.7 & 79.8 & 64.6 &
       74.1 & 74.6 & 65.2 & 55.1 & 81.0 &
       74.6 & 59.7 & 84.3 & 70.4 \\
     SRDC\cite{tang_unsupervised_2020} & 52.3 & 76.3 & 81.0 & 69.5 &
       76.2 & 78.0 & 68.7 & 53.8 & 81.7 &
       76.3 & 57.1 & 85.0 & 71.3 \\
     FixBi\cite{na_fixbi_2021} & 58.1 & 77.3 & 80.4 & 67.7 &
       79.5 & 78.1 & 65.8 & 57.9 &
       81.7 & 76.4 & 62.9 & 86.7 & 72.7 \\
     \hline
     AVATAR & \textbf{75.0} $\pm$0.6 & \textbf{82.7} $\pm$0.1& \textbf{83.2} $\pm$0.9 & \textbf{79.4} $\pm$0.5 & \textbf{84.2} $\pm$0.5 & \textbf{82.5}$\pm$1.0 & \textbf{74.1} $\pm$0.6 & \textbf{74.1} $\pm$0.6& \textbf{82.4}$\pm$0.6 & \textbf{80.5}$\pm$0.3 & \textbf{80.5} $\pm$0.9 & \textbf{85.7} $\pm$0.2 & \textbf{80.3} $\pm$0.1 \\
     \hline
     \end{tabular}
}
\end{center}
\label{table:2}
\end{table*}

\begin{table*}[!h]
\caption{Accuracy$(\%)$ on imageCLEF for unsupervised domain adaptation (ResNet-50). The best accuracy is indicated in bold. \* Reproduced by \cite{gu_spherical_2020}}
\begin{center}
\resizebox{\textwidth}{!}{
\begin{tabular}{|c|c|c|c|c|c|c|c|}
\hline
Method    & I$\rightarrow$P & P$\rightarrow$I & I$\rightarrow$C & C$\rightarrow$I & C$\rightarrow$P & P$\rightarrow$C & Avg           \\
\hline
\hline
ResNet-50 \cite{he_deep_2016} & 74.8            & 83.9            & 91.5            & 78.0            & 65.5            & 91.2            & 80.7          \\
DANN\cite{long_learning_2015}      & 75.0            & 86.0            & 96.2            & 87.0            & 74.3            & 91.5            & 85.0          \\
iCAN \cite{zhang_collaborative_2018}      & 79.5            & 83.9            & 91.5            & 78.0            & 65.5            & 91.2            & 87.4          \\
CDAN\cite{long_conditional_2018}      & 77.7            & 90.7            & 97.7            & 91.3            & 74.2            & 94.3            & 87.7          \\
MSTN*\cite{xie_learning_2018}      & 77.3            & 91.3            & 96.8            & 91.2            & 77.7            & 95.0            & 88.2          \\
SAFN+ENT\cite{xu_larger_2019}  & 79.3            & 93.3            & 96.3            & 91.7            & 77.6            & 95.3            & 88.9          \\
SymNets\cite{zhang_domain-symmetric_2019}   & 80.2            & 93.6            & 97.0            & 93.4            & 78.7            & 96.4            & 89.9          \\
RSDA-MSTN \cite{gu_spherical_2020}& 79.8            & \textbf{94.5}   & \textbf{98.0}   & \textbf{94.2}   & 79.2            & \textbf{97.3}   & \textbf{90.5} \\
\hline
AVATAR    & \textbf{89.4} $\pm$0.3   & 89.7$\pm$0.0           & 93.1$\pm$0.2            & 89.4$\pm$0.3            & \textbf{82.5}$\pm$0.9   & 92.4$\pm$0.1            & 89.4 $\pm$0.1 \\
\hline
\end{tabular}

}
\label{table:CLEF}
\end{center}
\end{table*}

\subsection{Comparison with State-Of-The-Art}
We report the results of Office-31, Office-Home, and ImageCLEF in Tables \ref{table:1} ,\ref{table:2}, and \ref{table:CLEF}, where target classification performances of other state-of-the-art methods are quoted from respective papers. Experiments are conducted three times for all three benchmarks, and the mean accuracy for each task is reported along with its standard deviation. 

For Office-31, AVATAR significantly improves the classification performance on complex adaptation tasks where the baseline ResNet-50 performance is low (e.g., D$\rightarrow$A and W$\rightarrow$A). Additionally, the overall performance of AVATAR is superior to that of the latest works, such as SRDC \cite{tang_unsupervised_2020} and FixBi \cite{na_fixbi_2021}, indicating the effectiveness of the proposed framework.

For Office-Home, AVATAR shows strong performance in all tasks. Furthermore, our method achieves an average accuracy of 80.3\%, significantly outperforming the state-of-the-art performance of FixBi \cite{na_fixbi_2021}. This result suggests that the proposed algorithm is promising for solving complex unsupervised adaptation problems.

For ImageCLEF, AVATAR achieves overall competitive performance on average against other SOTA models. In addition, it achieves the highest performance on more complex adaptation tasks, such as I$\rightarrow$P and C$\rightarrow$P, where the source-only-trained ResNet-50 model's performance is relatively low compared to other tasks.

Despite not always achieving the best performance, AVATAR offers several benefits compared to other state-of-the-art methods. First, AVATAR effectively leverages the advantages of both adversarial learning and self-supervised learning, enabling the model to achieve higher accuracy in various adaptation tasks. This is evident in the results obtained for Office-31 and Office-Home datasets, where AVATAR consistently outperforms other state-of-the-art methods. Second, AVATAR's framework is more robust to variations in the quality of the pseudo-labels, allowing the model to maintain high performance even against complex adaptation tasks. This is particularly demonstrated by the results using ImageCLEF, where our model performance is the highest in complex tasks such as I$\rightarrow$P and C$\rightarrow$P.

\subsection{Ablation studies}
To further validate the effectiveness of AVATAR, we examine the improvement in accuracy through ablation studies using Office-31 and Office-Home. Specifically, we evaluate the performance of the following variants of AVATAR:
\begin{itemize}
    \item[1.] Source model (discriminative source loss)
    \item[2.] Variant 1 (domain adversarial learning, discriminative source loss)
    \item[3.] Variant 2 (domain adversarial learning, discriminative source loss, discriminative target loss)
    \item[4.] AVATAR (full model)

\end{itemize}
The Source Model is trained using the discriminative source loss (Eq. \eqref{eq10}), which means only the source domain dataset is used for training. Variant 1 incorporates our proposed domain adversarial learning losses (Eq. \eqref{eq3}, \eqref{eq4}) into the Source Model. The discriminative target loss (Eq. \eqref{eq8}) is then added to create Variant 2. Finally, our sample selection strategy for the target domain (Eq. \eqref{eq9}) instead of the discriminative target loss to construct the AVATAR algorithm. The results are presented in Tables \ref{table:3} and \ref{table:4}. Our analysis demonstrates that each component of the proposed AVATAR algorithm contributes to the overall performance improvement. Moreover, utilizing our sample selection strategy increases performance, particularly for challenging tasks.

\begin{table*}[!h]
 \caption{
 Accuracy$(\%)$ on Office-31 for the ablation study. The definition of each variant is explained in the main text. The best accuracy is indicated in bold. }
\begin{center}
\resizebox{\textwidth}{!}{
\begin{tabular}{|c|c|c|c|c|c|c|c|c|c|c|c|c|c|}
 \hline
 Method & $L_{dis}^s$ & $L_{dis}^t$ & $L_{adv_f}$ & $L_{adv_g}$ & Conf & A $\rightarrow$ W & D $\rightarrow$ W & W $\rightarrow$ D & A $\rightarrow$ D & D $\rightarrow$ A & W $\rightarrow$ A & Avg \\ [0.5ex]
 \hline\hline
 Source Model & \checkmark & & & & & 77.3 & 91.8 & 97.6 & 80.7 & 56.9 & 60.7 & 63.5\\
 Variant 1 & \checkmark & & \checkmark & \checkmark & & 89.4 & \textbf{98.2} & \textbf{100.0} & 85.9 & 66.1 & 66.6 & 84.4\\
 Variant 2 & \checkmark & \checkmark & \checkmark & \checkmark & & 94.1 & 98.1 & \textbf{100.0} & 90.4 & 74.0 & 72.8 & 88.2\\
 AVATAR & \checkmark & \checkmark & \checkmark & \checkmark & \checkmark & \textbf{95.0} & 98.0 & \textbf{100.0} & \textbf{92.3} & \textbf{86.4} & \textbf{85.3} & \textbf{92.8}\\
 
 \hline
 \end{tabular}%
 }
 \end{center}
\label{table:3}
\end{table*}

\begin{table*}[!h]
\caption{
 Accuracy$(\%)$ on Office-Home for the ablation study. The definition of each variant is explained in the main text. The best accuracy is indicated in bold.}
\begin{center}
\resizebox{1.0\textwidth}{!}{
     \begin{tabular}{|c|c|c|c|c|c|c|c|c|c|c|c|c|c|}
     \hline
      Method& Ar$\rightarrow$Cl & Ar$\rightarrow$Pr & Ar$\rightarrow$Rw & Cl$\rightarrow$Ar & Cl$\rightarrow$Pr & Cl$\rightarrow$Rw & Pr$\rightarrow$Ar & Pr$\rightarrow$Cl & Pr$\rightarrow$Rw & Rw$\rightarrow$Ar & Rw$\rightarrow$Cl & Rw$\rightarrow$Pr & Avg \\ [0.5ex]
     \hline\hline
     Source Model & 46.2 & 68.4 & 75.4 & 54.1 & 63.9 & 66.9 & 55.3 & 40.9 & 74.7 & 66.9 & 47.3 & 79.3 & 61.6 \\
     Variant 1 & 54.8 & 69.9 & 76.0 & 60.7 & 68.8 & 69.7 & 60.0 & 56.4 & 76.2 & 72.7 & 59.9 & 82.7 & 67.3 \\
     Variant 2 & 57.5 & 74.7 & 77.7 & 65.0 & 74.6 & 73.7 & 67.0 & 61.1 & 79.0 & 74.9 & 63.4 & 84.7 & 71.1 \\
      AVATAR & \textbf{74.4} & \textbf{82.7} & \textbf{82.3} & \textbf{79.6} & \textbf{83.7} & \textbf{83.0} & \textbf{74.2} & \textbf{73.2} & \textbf{81.6} & \textbf{80.5} & \textbf{81.2} & \textbf{85.4} & \textbf{80.1} \\
     \hline
     \end{tabular}
     }
 \end{center}
\label{table:4}
\end{table*}

\subsection{Feature visualization} 
\begin{figure*}[!h]
\centering
\includegraphics[width=\textwidth]{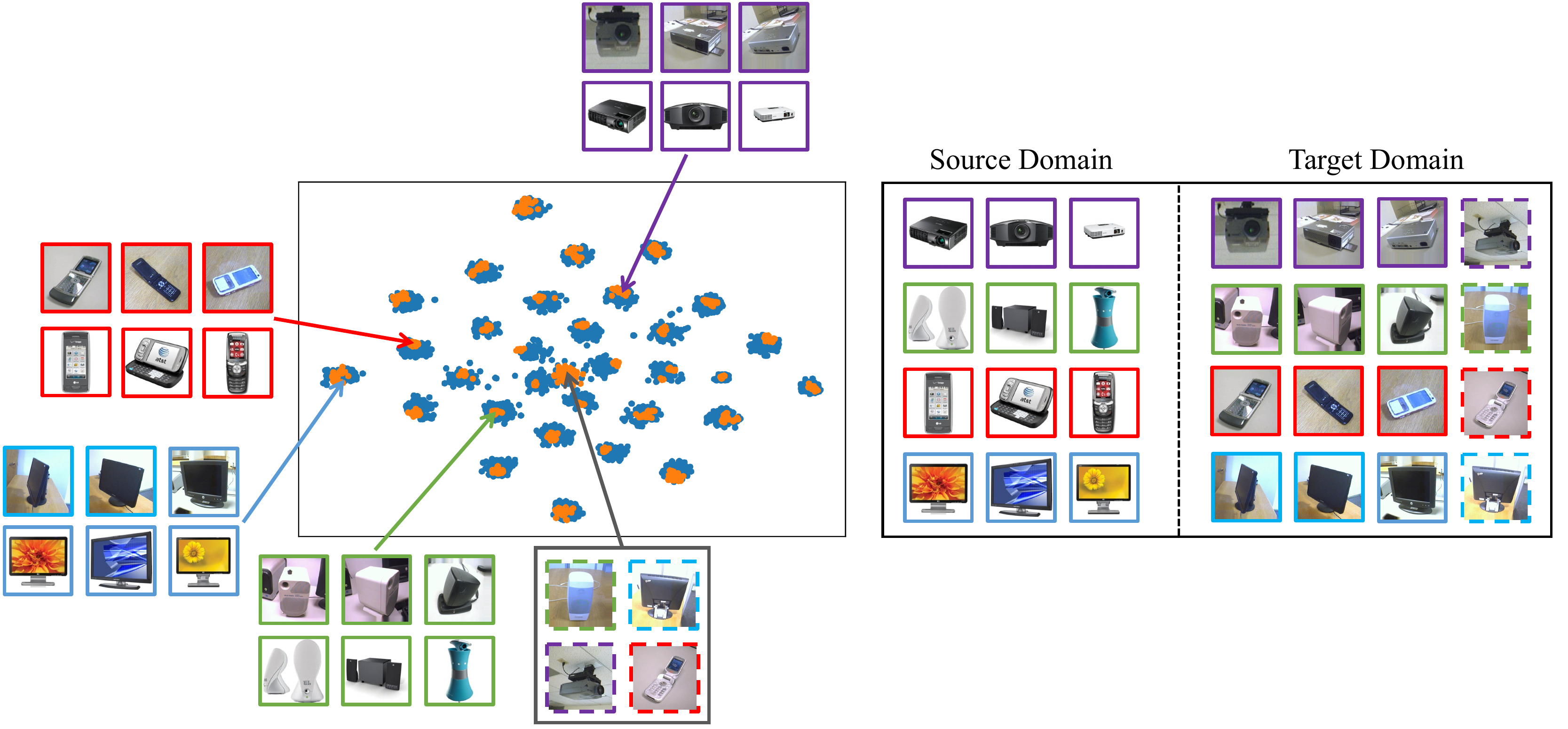}
\caption{t-SNE feature embedding visualization on $\mathcal{Z}$ using the target domain data for A $\rightarrow$ W. Note that source domain embedding is denoted in blue, and target domain embedding is denoted in orange. Best viewed in color.}
\label{fig:2}
\end{figure*}

In Figure \ref{fig:2}, we visualize the Office-31 source and target features in $\mathcal{Z}$ for task A$\rightarrow$W using t-SNE embedding \cite{van_der_maaten_visualizing_2008}. The visualization shows that most of the target domain features are embedded along with the source domain features to form discriminative clusters, indicating that these target instances are correctly classified and assigned to the same cluster as the source domain. However, some target samples are moved towards the center and regarded as difficult-to-classify target samples in our sample selection strategy for the target domain. 
The visualization suggests that AVATAR moves negative samples toward the outside of corresponding clustering decision boundaries, maintaining a high level of discrimination for each cluster. This results in well-separated clusters with high purity.

\subsection{Analysis of confidence threshold for the sample selection strategy} 
\begin{figure*}[!h]
\centering
\includegraphics[width=\textwidth]{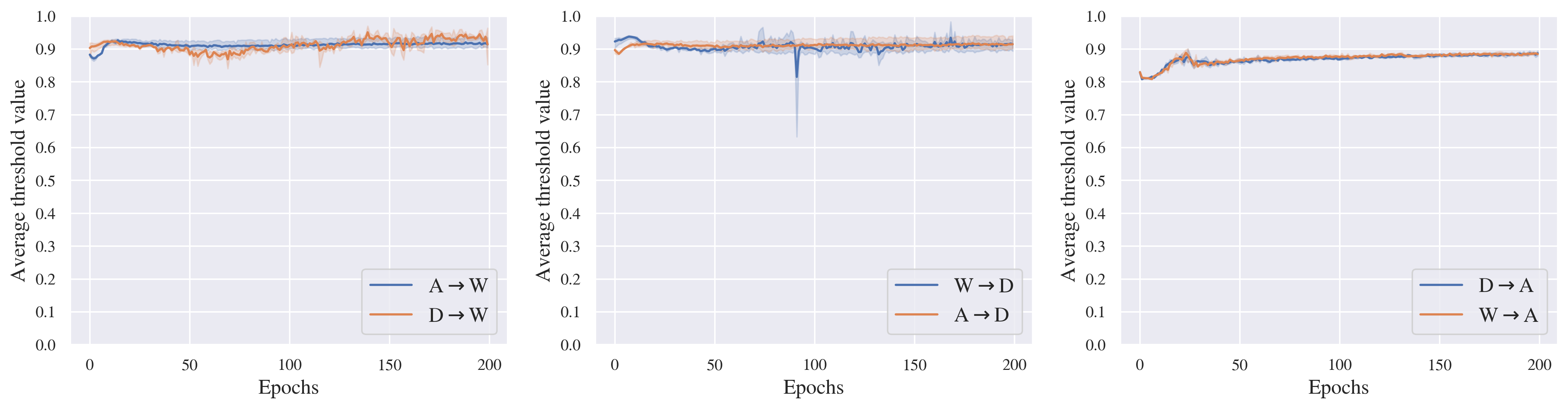}%
\caption{
Evolution of threshold values over 200 training epochs on the Office-31 dataset. The line segments represent the average threshold values for all 31 classes, and the shaded segments represent the 95\% confidence interval obtained from three independent experiments. Best viewed in color.}
\label{fig:3}
\end{figure*}

\begin{figure*}[!h]
\centering
\includegraphics[width=\textwidth]{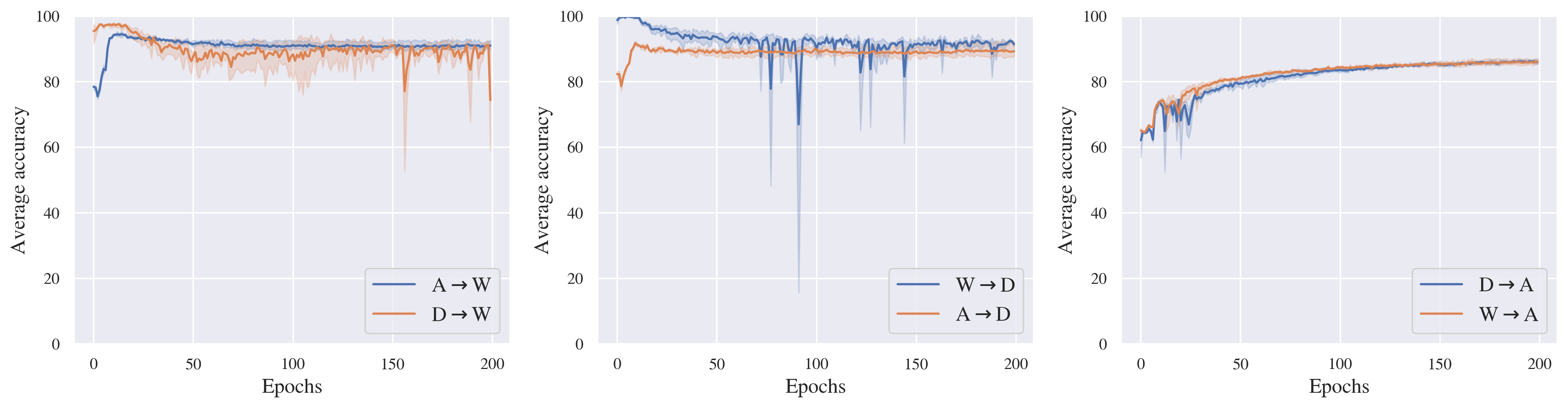}%
\caption{Transition of accuracy through 200 epochs using Office-31. Line segments show the average accuracy of 31 classes. Shaded segments show the 95\% interval using three experiments. Best viewed in color.}
\label{fig:4}
\end{figure*}

In Figure \ref{fig:3} and Figure \ref{fig:4}, we compare the average values of $\{\tau\}_{k=1}^K$ and accuracy for different tasks in Office-31. Note that $\{\tau\}_{k=1}^K$ are changed adaptively in each epoch, and those values are stable in most of the tasks, ranging between 0.80 and 0.95, except for two of the easiest tasks (i.e., D$\rightarrow$W and W$\rightarrow$D). We also observe a higher variance in accuracy in these tasks, whereas accuracy in other tasks stabilizes at the end of training. In addition, we observe a decrease in accuracy for these easiest tasks and a significant improvement of accuracy in difficult tasks (e.g., W$\rightarrow$A and D$\rightarrow$A) during training. These observations suggest that our sample selection strategy for the target domain is less effective when the task is easier (i.e., most target samples are easy to discriminate).

Nonetheless, we still observe an improvement in performance at the early stage of training for these more manageable tasks. Based on these observations, we conclude that AVATAR training should be terminated earlier when the UDA task is relatively easy. On the other hand, full training of 200 epochs is recommended for complex UDA tasks.

\subsection{Toward solving imbalanced UDA tasks} 
To assess the effectiveness and resilience of our AVATAR model in addressing imbalanced unsupervised domain adaptation (UDA) tasks, we conducted an experiment using the VISDA-2017 dataset. This dataset is an extensive and comprehensive dataset for synthetic-to-real domain adaptation, featuring a vast collection of images, including 152,397 synthetic images for the source domain and 55,388 real-world images for the target domain. The source domain consists of labeled images for 12 object categories, including vehicles, traffic lights, and people. The target domain features images captured from urban environments and annotated with the same set of labels for testing.

To evaluate the performance of our proposed approach on realistic imbalanced UDA tasks, we downsampled the source domains by 90\% to decrease the source domain sample size, making the adaptation task more challenging. Furthermore, we deliberately introduced class imbalance for the target domain dataset by using the airplane as the majority class and 14 other classes as minority classes. We then downsampled the minority classes based on the number of samples in the majority class, gradually reducing the number of samples in the minority classes from 10\% of the majority class to 1\% by 1\%. As a result, we created 10 imbalanced sub-datasets with varying levels of class imbalance. Finally, we trained the AVATAR model using the same approach described in Subsection \ref{subsection:setup} for all 10 sub-datasets, assessed its performance, and compared its performance against 3 variant models defined in our ablation study.

The results, as shown in Table \ref{table:minority}, demonstrate that our proposed model consistently perform better even when the target domain dataset is highly imbalanced. The differences between the AVATAR model's performance and the model trained using only the source domain dataset are significant. Our results suggest that the AVATAR model is robust against fewer samples in target domain minority classes, a common issue in real-world UDA scenarios. This experiment indicates the efficacy and adaptability of the AVATAR model in solving imbalanced UDA tasks.

\begin{table*}[!h]
\caption{Average accuracy$(\%)$ of minority classes for the target domain using VISDA-2017 dataset.
AVATAR is compared against Source Model, Variant 1, and Variant 2 from the ablation study. Best accuracy is indicated in bold.}

\begin{center}
\resizebox{1.0\textwidth}{!}{
\begin{tabular}{|c|c|c|c|c|c|c|c|c|c|c|}
\hline\
Minority class sample size (\%) & 1.0  & 2.0  & 3.0  & 4.0  & 5.0  & 6.0  & 7.0  & 8.0  & 9.0  & 10.0 \\
\hline
\hline
Source Model                         & 53.6 & 52.0 & 50.6 & 50.4 & 50.6 & 50.4 & 50.7 & 50.9 & 50.9 & 51.0 \\
Variant 1 & 53.6 & 53.4 & 57.1 & 61.7 & 61.4 & 63.0 & 60.5 & 60.1 & 60.1 & 63.2 \\
Variant 2 & 53.6 & 55.0 & 56.2 & 58.3 & 61.2 & 61.3 & 60.8 & 62.7 & 63.2 & 62.7 \\
AVATAR & \textbf{58.1} & \textbf{58.2} & \textbf{62.5} & \textbf{67.3} & \textbf{63.8} & \textbf{69.0} & \textbf{69.0} & \textbf{74.3} & \textbf{75.5} & \textbf{71.7} \\
\hline
\end{tabular}
     }
 \end{center}
\label{table:minority}
\end{table*}

\section{Conclusion}
In this study, we proposed AVATAR, an \textbf{A}dversarial self-supervised domain \textbf{V}ariant \textbf{A}daptation network for \textbf{TAR}get domain, which effectively improves discrimination performance by combining weighted adversarial learning, self-supervised learning, and robust sample selection strategy for the target domain, all guided by deep clustering. Our performance comparisons against various state-of-the-art methods using standard UDA benchmarks demonstrate that AVATAR outperforms other methods, particularly for complex UDA tasks where the accuracy of the baseline model is low. In addition, extensive ablation studies and experiments validate the effectiveness of the proposed algorithm. While this paper focuses on the problem of unsupervised domain adaptation, our approach can also be applied in a semi-supervised setting. Our future works will evaluate the effectiveness of our proposed model in this scenario. Also, our proposed model can be extended using other network architectures as its backbones, including ResNet101 \cite{he_deep_2016} and ViT \cite{dosovitskiy_image_2021}, and we will investigate the performance of AVATAR on these networks.


\bibliographystyle{elsarticle-harv} 
\bibliography{informationScience_submission}

\begin{algorithm}
\caption{Training procedure for AVATAR. $E$ denotes the training epoch. $i$ denotes the training iteration. $B_s$ and $B_t$ denote mini-batches for the source and target domains, respectively.}
\begin{algorithmic}[1]
\small
\floatname{algorithm}{Procedure}

\renewcommand{\algorithmicrequire}{\textbf{Input:}}
\renewcommand{\algorithmicensure}{\textbf{Output:}}
  \REQUIRE Labelled data from source domain $\mathcal{X}^s = \{(x_i^s,y_i^s)\}_{i=1}^{N^s}$; unlabeled target domain data $\mathcal{X}^t = \{(x_i^t)\}_{i=1}^{N^t}$; warm-up epoch size $e$; regularization parameter $\lambda$
  \ENSURE $\theta$, $\vartheta$
  
  \STATE Initialization: $q_{i, k}^t = I[k=\hat{y}_i^{t}]$ for $i \in \{ 1, \dots, N^t \}$ and $k \in \{1, \dots, K\}$, $\{w_i^{s}\}_{i=1}^{N^s}=\{w_i^{t}\}_{i=1}^{N^t} = 1$, $\{\tau_k^t \}_{k=1}^K = 0$, $E = 1$

  \WHILE{not converged}
    \STATE Update $w_i^s, i \in \{1,\dots,N^s\}$ by Eq.\eqref{eq1}
    \STATE Update $w_i^t, i \in \{1,\dots, N^t\}$ by Eq.\eqref{eq2}
    \STATE Update $\{\mathbf{c}_{k}^{t} \}_{k=1}^{K}$ by kernel K-means
    \IF{$E > e$}
    \STATE Update $\bar{w}_k^t, k \in \{1,\dots,K \}$ by using Eq.\eqref{eq7}
    \STATE Update $\tau_k^t, k \in \{1,\dots,K \}$ by using Eq.\eqref{eq8}
    \ENDIF
    \FOR{$i=1$ to MAXITER}
        \STATE Sample $B_s$ and $B_t$ from $\mathcal{X}^s$ and $\mathcal{X}^t$
        \STATE Update $q_{i, k}^t$ based on Eq.\eqref{eq5}
        \STATE Update $\theta$ and $\vartheta$ by minimizing Eq.\eqref{eq12} and Eq.\eqref{eq13} on $B_s$ and $B_t$
        \ENDFOR
        \STATE $E = E + 1$
 \ENDWHILE

\label{alg:the_alg}
\end{algorithmic}
\end{algorithm}

\end{document}